\documentclass[lettersize,journal]{IEEEtran}
\usepackage{amsmath,amsfonts}
\usepackage{algorithmic}
\usepackage{algorithm}
\usepackage{array}
\usepackage[caption=false,font=normalsize,labelfont=sf,textfont=sf]{subfig}
\usepackage{textcomp}
\usepackage{stfloats}
\usepackage{url}
\usepackage{verbatim}
\usepackage{graphicx}
\usepackage{hyperref}
\usepackage{color}
\usepackage{xspace}
\usepackage{setspace}%zhuy added
\usepackage[numbers]{natbib}%zhuy added
\usepackage{multirow}%zhuy add
\usepackage{amssymb} %zhuy add
\usepackage{multirow}
\usepackage{todonotes}
\usepackage{xcolor}
\newif{\ifhidecomments}
\hidecommentsfalse
\ifhidecomments
\newcommand{\wjdd}[1]{\todo[linecolor=cyan,backgroundcolor=cyan!25,bordercolor=cyan,size=\scriptsize]{}
\newcommand{\wjd}[1]{{\color{cyan}{}
\else
\newcommand{\wjdd}[1]{\todo[linecolor=cyan,backgroundcolor=cyan!25,bordercolor=cyan,size=\scriptsize]{(Jindong) #1}}
\newcommand{\wjd}[1]{{\color{cyan}{[(Jindong) #1]}}}
\fi

\newcommand{\prompt}[1]{{\small \ttfamily #1}}
\newcommand{\method}{SAFE\xspace}
\newcommand{\methoda}{SAFE-A\xspace}
\begin{document}

\title{Enhancing Few-shot CLIP with Semantic-Aware Fine-Tuning}

\author{Yao Zhu, Yuefeng Chen, Wei Wang, Xiaofeng Mao, Xiu Yan, Yue Wang, Zhigang Li, Wang lu, Jindong Wang, \\ Xiangyang Ji
        % <-this % stops a space
\thanks{Yao Zhu, Xiangyang Ji, Zhigang Li, Yue Wang, Wang Lu are with the Tsinghua University, Beijing, China, 100084.
(E-mail: ee$\_$zhuy$@$zju.edu.cn, \protect\\ 
xyji@tsinghua.edu.cn,lzg.matrix@gmail.com,wangyue.2017@tsinghua.org.cn, luw12@tsinghua.org.cn)}

\thanks{Yuefeng Chen, Xiaofeng Mao are with the Alibaba Group, Hangzhou,
China, 310023. (E-mail: 
yuefeng.chenyf@alibaba-inc.com, mxf164419@alibaba-inc.com)}

\thanks{Xiu Yan is with the Meituan Group, Beijing,
China, 100102. (E-mail: 
yanx18@tsinghua.org.cn)}

\thanks{Wei Wang is with the Dalian University of Technology-Ritsumeikan University (DUT-RU) International School of Information Science \& Engineering, Dalian University of Technology, Dalian, Liaoning, 116620, China. (E-mail: WWLoveTransfer@mail.dlut.edu.cn)}
\thanks{Jindong Wang is with the Microsoft Research Asia, Beijing,
China, 100080. (E-mail: 
jindong.wang@microsoft.com)}

\thanks{Corresponding author: Wang Lu.}
% \thanks{Manuscript received April 19, 2021; revised August 16, 2021.}
}

\maketitle

\begin{abstract}
Learning generalized representations from limited training samples is crucial for applying deep neural networks in low-resource scenarios. Recently, methods based on Contrastive Language-Image Pre-training (CLIP) have exhibited promising performance in few-shot adaptation tasks. To avoid catastrophic forgetting and overfitting caused by few-shot fine-tuning, existing works usually freeze the parameters of CLIP pre-trained on large-scale datasets, overlooking the possibility that some parameters might not be suitable for downstream tasks. To this end, we revisit CLIP's visual encoder with a specific focus on its distinctive attention pooling layer, which performs a spatial weighted-sum of the dense feature maps. Given that dense feature maps contain meaningful semantic information, and different semantics hold varying importance for diverse downstream tasks (such as prioritizing semantics like ears and eyes in pet classification tasks rather than side mirrors), using the same weighted-sum operation for dense features across different few-shot tasks might not be appropriate.
Hence, we propose fine-tuning the parameters of the attention pooling layer during the training process to encourage the model to focus on task-specific semantics.
In the inference process, we perform residual blending between the features pooled by the fine-tuned and the original attention pooling layers to incorporate both the few-shot knowledge and the pre-trained CLIP's prior knowledge. We term this method as \textbf{S}emantic-\textbf{A}ware \textbf{F}in\textbf{E}-tuning (\method).
\method is effective in enhancing the conventional few-shot CLIP and is compatible with the existing adapter approach (termed \methoda).
Extensive experiments on 11 benchmarks demonstrate that both \method and \methoda significantly outperform the second-best method by +1.51$\%$ and +2.38$\%$ in the 1-shot setting and by +0.48$\%$ and +1.37$\%$ in the 4-shot setting, respectively.
\end{abstract}

\begin{IEEEkeywords}
Vision-language learning, few-shot classification, fine-tuning.
\end{IEEEkeywords}

\section{Introduction}

\IEEEPARstart{D}{ue} to the advancements in model architecture and the availability of large-scale, high-quality training datasets, visual understanding tasks such as image classification \cite{li2023imagenet,hsu2023abc}, semantic segmentation \cite{bi2023interactive,bai2022human,lang2023retain}, and object detection \cite{mao2023coco,cong2022cir} have shown significant progress in recent years. However, collecting large-scale, high-quality datasets in real-world scenarios is costly and resource-intensive.  
Thanks to the Contrastive Language-Image Pre-training model (CLIP)~\cite{radford2021CLIP}, which demonstrates outstanding capability in aligning vision-language representations by learning from 400 million internet-sourced image-text pairs, users can craft zero-shot classifiers with favorable performance by constructing suitable prompt. For instance, the confidence score of an image belonging to a new ``\prompt{[CLASS]}" in open-vocabulary settings can be obtained by using the CLIP's similarity measure between the image and the prompt sentence ``a photo of a \prompt{[CLASS]}" without further re-training.

% \begin{figure}[tbh!]
% \includegraphics[width=0.48\textwidth]{Average.pdf}
% \centering
% \caption{Comparison with existing few-shot adaptation methods of leveraging CLIP. We report the average results on 11 downstream datasets with various training samples. Our methods, SAFE and SAFE-A, consistently and substantially outperform the existing approaches.}
% \label{Fig:average}
% \end{figure}

However, zero-shot CLIP's performance falls short in some scenarios. For example, on the EuroSAT dataset \cite{helber2019eurosat}, its classification accuracy with the ResNet-50 visual encoder is a mere 37.56$\%$. To enhance the practicality and applicability of CLIP, few-shot learning leverages a small number of samples to enhance CLIP's adaptability in downstream tasks. This holds significant importance for the application of deep learning techniques in low-resource scenarios.
Following the direction of prompt tuning \cite{liu2023pre}, CoOp \cite{zhou2022learning} proposes to learn continuous textual tokens via minimizing the classification loss on the target task, and CoCoOp \cite{zhou2022conditional} extends CoOp by further learning an intermediate neural network to generate an input-conditional token for each image. Inspired by the parameter-efficient fine-tuning techniques \cite{houlsby2019parameter} used in natural language processing, CLIP-Adapter \cite{gao2023clipadapter} equips CLIP with two learnable residual adapters and TIP-Adapter \cite{zhang2022tipadapter} constructs a key-value cache model initialized with CLIP-extracted features from few-shot samples. CLIP-Adapter and TIP-Adapter optimize the added lightweight linear layers while freezing CLIP's visual encoder. 
These existing methods share a common characteristic of freezing the CLIP's parameters pre-trained on large-scale datasets to avoid overfitting caused by few-shot fine-tuning \cite{serra2018overcoming,wortsman2022robust,gao2022bmu}.
However, are all the pre-trained parameters of the visual encoder suitable for downstream tasks, and is it reasonable to freeze all these parameters?
% and have indeed achieved good performance on few-shot tasks. The question arises: \textit{how can we further enhance the performance of few-shot CLIP?}
% \wjdd{This motivation is weak. You should clearly say the shortcomings of them: adapter method requires full understanding of the CLIP inner structure; free-fine tuning may lost the intrinsic information of CLIP.} 

\begin{figure*}[tbh!]
\includegraphics[width=0.97\textwidth]{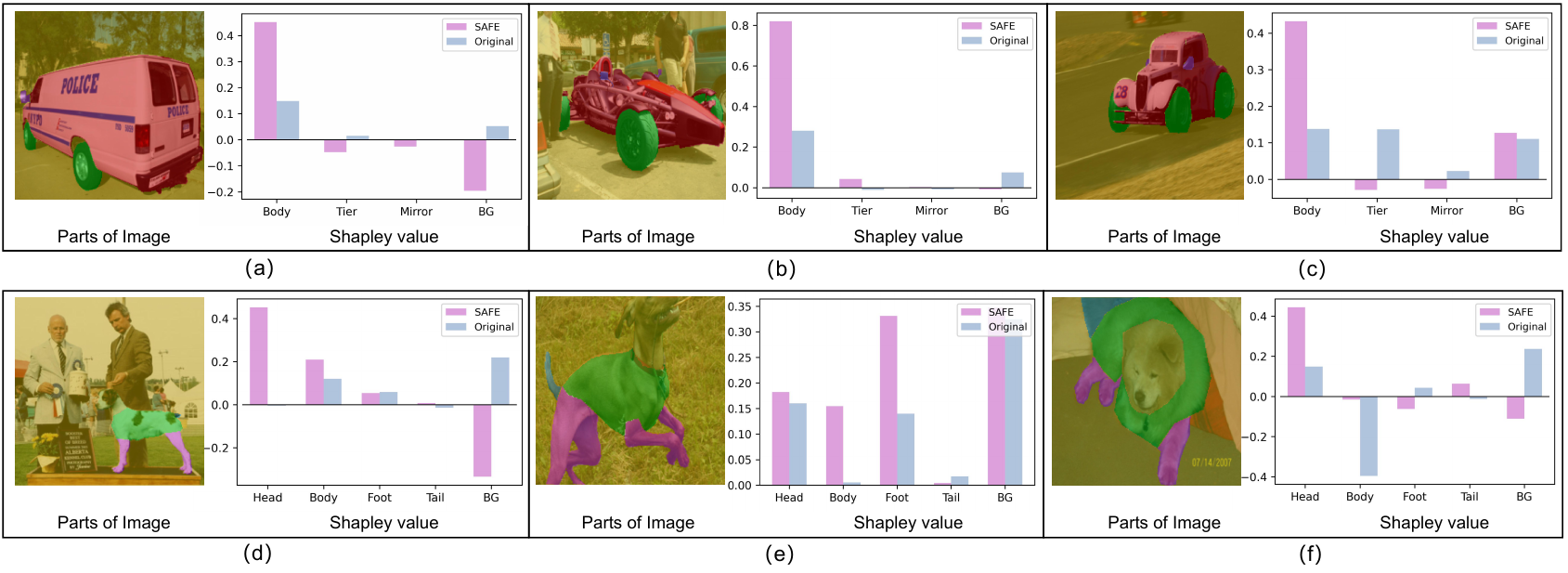}
\centering
\caption{We use the PartImageNet \cite{he2022partimagenet} segmentations to compare the Shapley value of each part when using the original CLIP and the CLIP fine-tuned by our SAFE method. A dog is divided into four parts: 'head', 'body', 'leg', and 'tail', while a car is divided into three parts: 'body', 'tire', and 'side mirror'. We refer to the regions outside the objects as the background (denoted as BG). Different colors in the image represent different semantic parts. We use ResNet-50 as the visual backbone.}
\label{Fig:shapley}
\end{figure*}

We explored this question from the perspective of semantic attention and hypothesized that different semantics contribute differently to various downstream tasks. Specifically, features such as headlights, wheels, and side mirrors may hold significant importance in the car classification task, whereas these semantics might carry minimal relevance in the pet classification task. We introduce Shapley values \cite{lundberg2017unified,chen2023algorithms} to measure the importance of different parts of the image concerning the final probability of the ground truth class and find that the CLIP's visual encoder pre-trained on large-scale datasets does not always focus on task-specific semantics. As shown in Figure \ref{Fig:shapley} (d), the original CLIP model places more importance on the background region relative to the dog's semantics when making decisions. Therefore, adopting all the pre-trained parameters for the downstream task might not be an appropriate choice and we should explore ways to guide the model's attention toward task-specific semantics, which has the potential to enhance the model's interpretability and few-shot adaptation performance.

Upon examining the CLIP's dense features (before undergoing the pooling layer), we discovered an intriguing phenomenon: these dense features have captured semantic information about the objects, which can be used to establish semantic correspondences between various images, as discussed in Section \ref{Motivation}. In order to better utilize the semantic information within the dense features, this paper thoroughly revisits CLIP's visual encoder, especially its unique attention pooling layer. Compared to the conventional global average pooling, the attention pooling layer performs a spatially weighted sum of the dense features, which means that this layer assigns different weights to various locations in the dense features. In order to rectify the weights of different parts in the dense features according to their significance for downstream tasks, we propose using few-shot training samples to fine-tune CLIP's attention pooling layer. With the guidance of few-shot training samples, the attention pooling layer focuses more on the task-specific semantics within dense features, rather than adopting the same attention mechanism across different tasks as in existing methods \cite{zhang2022tipadapter,lin2023multimodality}.
During the inference process, we propose performing a residual blending between the output of the fine-tuned attention pooling layer and the original attention pooling layer, aiming at incorporating both the few-shot knowledge and the pre-trained CLIP's prior knowledge.
This approach can achieve both remarkable few-shot adaptation performance and out-of-distribution generalization performance. 
We term this method \textbf{S}emantic-\textbf{A}ware \textbf{F}in\textbf{E}-tuning (SAFE). It can also be plug-and-play utilized to enhance the existing method that equips CLIP with adapters to enhance the few-shot performance (denoted as SAFE-A). Taking the dog image in Fig. \ref{Fig:shapley}(f) as an example, the head part has a decisive impact on the decision of the CLIP fine-tuned by SAFE, while the tail and background have a slight influence on the decision. This aligns with human perception, where the head features are crucial for pet classification. SAFE not only endows few-shot CLIP with favorable classification performance but also enhances its interpretability.

\begin{figure}[tbh!]
\includegraphics[width=0.46\textwidth]{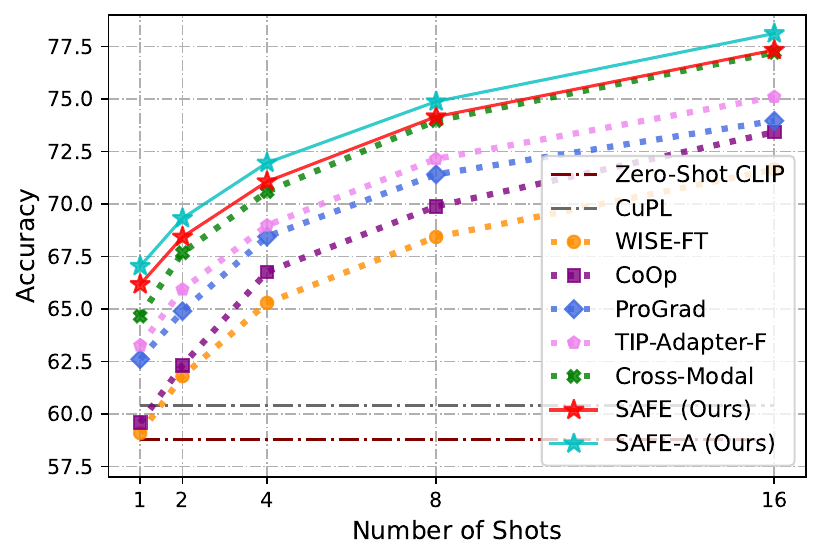}
\centering
\caption{Comparison with existing few-shot adaptation methods of leveraging CLIP. We report the average results on 11 downstream datasets with various training samples. Our methods, SAFE and SAFE-A, consistently and substantially outperform the existing approaches.}
\label{Fig:average}
\end{figure}

We conduct large-scale experiments on 11 widely used datasets to demonstrate the effectiveness of our method, which has a consistent and substantial improvement over existing baselines, as illustrated in \figurename~\ref{Fig:average}. For instance, SAFE and SAFE-A with 1 sample per category surpass the second-best method by +1.51$\%$ and +2.38$\%$, respectively, for the average accuracy. Furthermore, SAFE can achieve superior generalization capability on out-of-distribution datasets than existing methods.
The main contributions of our paper are summarized as follows:
 \begin{itemize}

 % \item[$\bullet$] We propose a semantic-aware fine-tuning (SAFE) method, which fine-tunes CLIP's attention pooling layer to encourage the model to focus on the task-specific semantics in dense features and blends the outputs of the fine-tuned and the original attention pooling layer to incorporate the few-shot knowledge and the pre-trained prior knowledge.
 
 \item[$\bullet$] We observed an intriguing phenomenon that CLIP's dense features have captured semantic information about objects and can be used to establish semantic correspondences between different images. We hypothesize that different semantic information holds varying importance for the downstream few-shot task.
 % We propose blending the outputs of the fine-tuned attention pooling layer and the original attention pooling layer in a residual manner, which aims to address the potential catastrophic forgetting caused by fine-tuning and further enhance the model's generalization capability.

 \item[$\bullet$] We propose a semantic-aware fine-tuning (SAFE) method, which fine-tunes CLIP's attention pooling layer to encourage the model to focus on the task-specific semantics in dense features and blends the outputs of the fine-tuned and the original attention pooling layer to incorporate the few-shot knowledge and the pre-trained prior knowledge.

 \item[$\bullet$] 
  Extensive experiments demonstrate that our proposed SAFE and SAFE-A achieve state-of-the-art performance on 11 widely adopted few-shot classification datasets and exhibit superior out-of-distribution generalization capabilities.
 \end{itemize}
 
The rest of the paper is organized as follows: Section \ref{sec:related} summarizes the literature related to few-shot CLIP. In Section \ref{sec:method}, we begin by outlining the motivation for our approach. Subsequently, we introduce SAFE (\textbf{S}emantic-\textbf{A}ware \textbf{F}in\textbf{E}-tuning), a method that fine-tunes the attention pooling layer to encourage the model to focus on task-specific semantics. Additionally, we explore how to utilize SAFE to enhance the existing adapter methods. Section \ref{sec:experiments} covers the experimental setup in detail. We then conduct experiments on 11 few-shot classification benchmarks and assess the out-of-distribution capability of our proposed SAFE and SAFE-A methods. After that, we provide an ablation discussion to further understand our method. Finally, Section \ref{sec:conclusion} presents the conclusive results.

\section{Related Work}\label{sec:related}

 With the assistance of large-scale and high-quality datasets, deep models demonstrate outstanding performance \cite{krizhevsky2012imagenet}. However, the collection of such data is challenging due to long-tail distributions, noisy annotations, and escalating labeling costs. To address this issue, few-shot learning has been proposed \cite{ji2023memorizing,zhou2023improving,vu2023instance,cheng2022imposing}. Classical few-shot learning algorithms usually assume a large-scale meta-training set for pre-training the deep model, followed by evaluation on multiple episodes of few-shot training (support) and test (query) sets. Metric learning methods \cite{bateni2020improved,sung2018learning} compute distances between the query samples and known category samples, determining the classification result for the query samples based on proximity to neighboring categories. Meta-learning methods \cite{finn2017model,sun2019meta,chen2021meta} propose to learn meta-knowledge from a multitude of prior tasks, utilizing past knowledge to guide models in adapting to novel domains rapidly. Graph neural network methods \cite{satorras2018few,kim2019edge} leverage graph structures to model intra-class similarities and inter-class dissimilarities of the samples and have shown promising performance in some few-shot tasks.

 In recent developments, Contrastive Language-Image Pre-training (CLIP) \cite{radford2021CLIP} has demonstrated that learning from contrastive vision-language pairs yields promising generalized features for zero-shot recognition over diverse datasets. Furthermore, several efforts have started to devise efficient strategies for its adaptation to downstream few-shot tasks. In this study, we follow the new few-shot evaluation protocol as implemented in recent research on few-shot CLIP: (1) the meta-training phase in classical few-shot learning is supplanted by pre-trained CLIP models, and (2) the test sets consist of the official test set of each dataset (hence not few-shot).
 
 \textbf{Prompt tuning.} The main concept behind prompt tuning is to formalize diverse NLP tasks as masked language modeling problems \cite{kenton2019bert,radford2021learning}, akin to the pre-training of language models. Central to the success of these approaches is the identification of the appropriate prompt, which constitutes a pivotal aspect of research in this domain. \citet{zhou2022learning} propose Context Optimization (CoOp) to learn continuous soft prompts by optimizing a set of learnable prompt tokens with few-shot samples for replacing the carefully chosen hard prompts. \citet{zhou2022conditional} propose Conditional Context Optimization (CoCoOp), which trains an intermediate network to generate input-conditional tokens. Compared to CoOp's static prompts, CoCoOp's dynamic prompts are instance-adaptive. 
 Considering that learning a single prompt overlooks the diversity of visual representations and might face challenges in capturing the various changes in visual content, \citet{lu2022prompt} propose learning a distribution of diverse prompts to enhance generalization capabilities. \citet{zhu2023prompt} propose the Prompt-aligned Gradient method to prevent prompt tuning from forgetting the general knowledge acquired from visual language models. This approach selectively updates the prompt whose gradient aligns with the general knowledge, ensuring the preservation of general knowledge.
 In addition, large language models (LLMs)  can also be used for designing prompts. For instance, \citet{pratt2023does} leverage GPT3 \cite{brown2020language} to generate descriptive sentences that contain important discriminating characteristics of the image categories, achieving promising performance with these customized prompts. 
 
 \textbf{Model fine-tuning.} CLIP can adopt linear probing \cite{radford2021CLIP,chen2020simple} and full-finetuning \cite{kirkpatrick2017overcoming,mao2022context,zhang2022glipv2} when transferring to downstream tasks. However, fine-tuning using a small number of samples may lead to overfitting on the training data and sacrifice the performance on the test data \cite{wang2020generalizing}.
 In natural language processing, \citet{houlsby2019parameter} propose an Adapter to insert learnable linear layers into each Transformer layer while freezing the backbone to achieve data-efficient fine-tuning for downstream tasks. Inspired by this approach, \citet{gao2023clipadapter} propose CLIP-Adapter, which conducts fine-tuning with feature adapters on either the visual or language branch. \citet{zhang2022tipadapter} introduce Tip-Adapter, which constructs the adapter using a key-value cache model derived from the few-shot training set. Compared to CLIP-Adapter, Tip-Adapter benefits from its superior initialization method, enabling faster convergence during fine-tuning. WiSE-FT \cite{wortsman2022robust} adopts a weight-space ensemble of CLIP's original pre-trained weights and fine-tuned weights to enhance out-of-distribution robustness. CALIP \cite{guo2023calip} proposes to guide the interaction between visual and textual representations and explore cross-modal informative features through a parameter-free attention mechanism. Its parametric counterpart CALIP-F \cite{guo2023calip} further achieves higher few-shot accuracy. Moreover, \citet{lin2023multimodality} propose a cross-modal few-shot learning method, repurposing textual features as training samples during fine-tuning. 
 
 Different from existing methods, we observed that CLIP's dense features before the pooling layer exhibit meaningful semantic properties. Therefore, we propose a semantic-aware fine-tuning approach that encourages the model to focus on distinct semantic information according to specific downstream tasks. Experimental results demonstrate the superiority of our method over existing fine-tuning methods.

 \section{Method} \label{sec:method}
 
In Section \ref{Motivation}, we first present the motivation of this study. Subsequently, in Section \ref{SAFE}, we propose to enhance the few-shot CLIP through a semantic-aware fine-tuning method.

\begin{figure*}[tbh!]
\includegraphics[width=0.9\textwidth]{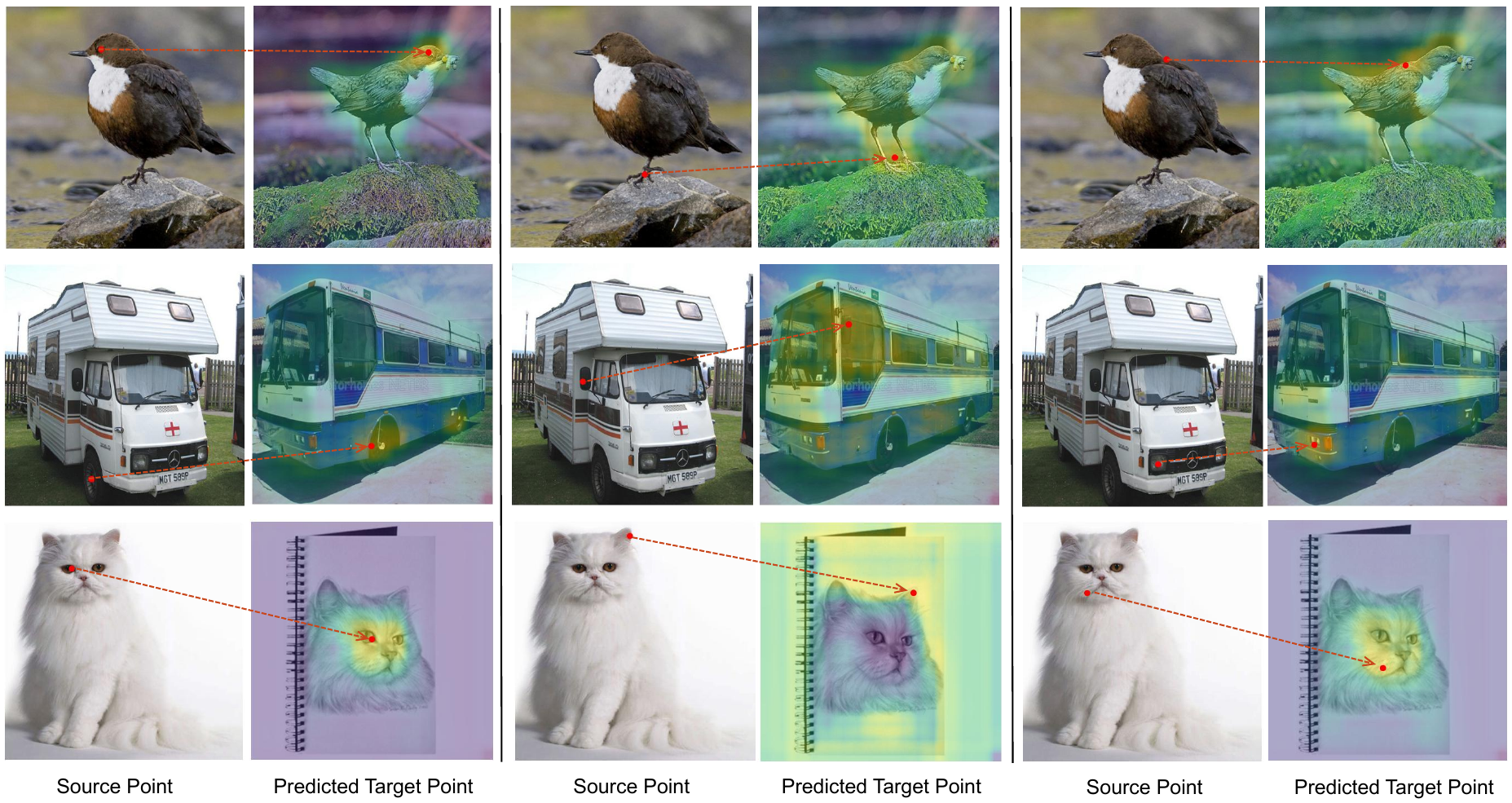}
\centering
\caption{The predicted target point given by CLIP's dense feature maps can somewhat semantically match the given red source point in the source image, which sheds light on the fact that the spatial region in CLIP's dense feature maps is relevant to the semantics of the object.}
\label{Fig:motivation}
\end{figure*}

\subsection{Motivation}\label{Motivation}
As humans, we are able to recognize a given new image by matching correspondence between the semantic features of the given image and our known categories. For instance, we use semantic features such as eyes, ears, and nose to determine whether an object is a cat. The question is, can Contrastive Language-Image Pre-training (CLIP) learn the semantic correspondence between images and leverage this information to enhance few-shot performance?
\citet{zeiler2014visualizing,natureindividual} show that the early layers of neural networks (closer to the input)  focus on local features such as edges, corners, and lines. Hence, we investigate whether the deep features (rather than early features) of the CLIP model exhibit generalized semantics that can be used to establish the correspondence between different images. 

Specifically, given a source image $\boldsymbol{x}_{s}$ and a source point $p_s$, we are interested in finding whether the dense features extracted by CLIP can be used to locate the semantically relevant point $p_t$ in the target image $\boldsymbol{x}_{t}$. Here, dense features refer to the features extracted by CLIP's visual feature extractor without undergoing the pooling process. We conduct bilinear interpolation upsampling on the dense feature maps to match their size with that of the input images and denote these upsampled dense features of $\boldsymbol{x}_{s}$ as $\boldsymbol{F}_{\boldsymbol{x}_{s}}$. The pixel-level feature at location $p_s$ is denoted as $\boldsymbol{F}_{\boldsymbol{x}_{s}} (p_s)$. Then we calculate the cosine similarity between $\boldsymbol{F}_{\boldsymbol{x}_{s}} (p_s)$ and different pixel-level  features of the target image and regard the most similar point as the predicted target point:
\begin{equation}
    p_t = \underset{p}{\text{argmax}} ~ \cos(\boldsymbol{F}_{\boldsymbol{x}_{s}} (p_s),\boldsymbol{F}_{\boldsymbol{x}_{t}} (p)),  \label{motivation}
\end{equation}
where $\cos(\cdot, \cdot)$ indicates the cosine similarity metric. 

As illustrated in Fig.~\ref{Fig:motivation}, given a specific spatial location on the source image (denoted as the source point), the feature maps extracted by CLIP can be utilized to identify the spatial location on the target image (referred to as the predicted target point) that is semantically related to the source point. We use the heatmap to represent the cosine similarity between points in the target image and the source point in the source image. Higher brightness indicates higher similarity, and vice versa. The red point in the source image marks the source point, while the point in the target image with the highest similarity is regarded as the predicted target point and is marked in red. It is evident that CLIP's dense features have captured nontrivial semantic information for visual recognition, such as the beak and claws of a bird, the headlights and tires of a bus, and the eyes and ears of a cat, which inspires us to leverage this information to enhance the model's few-shot performance.

We hypothesize that different downstream tasks might focus on distinct semantic parts of the object. For example, in a car classification task, the focus could be on semantic features like wheels, side mirrors, and headlights, while in a pet classification task, the focus might be on eyes, ears, and tails. Therefore, leveraging limited training samples to guide the model's attention towards task-specific semantics in CLIP's dense features might be an effective approach to enhance few-shot adaptation performance. However, existing methods always freeze the visual encoder and apply the same attention mechanism to dense features across different tasks, overlooking the need for task-specific adaptation.

\begin{figure*}[tbh!]
\includegraphics[width=0.825\textwidth]{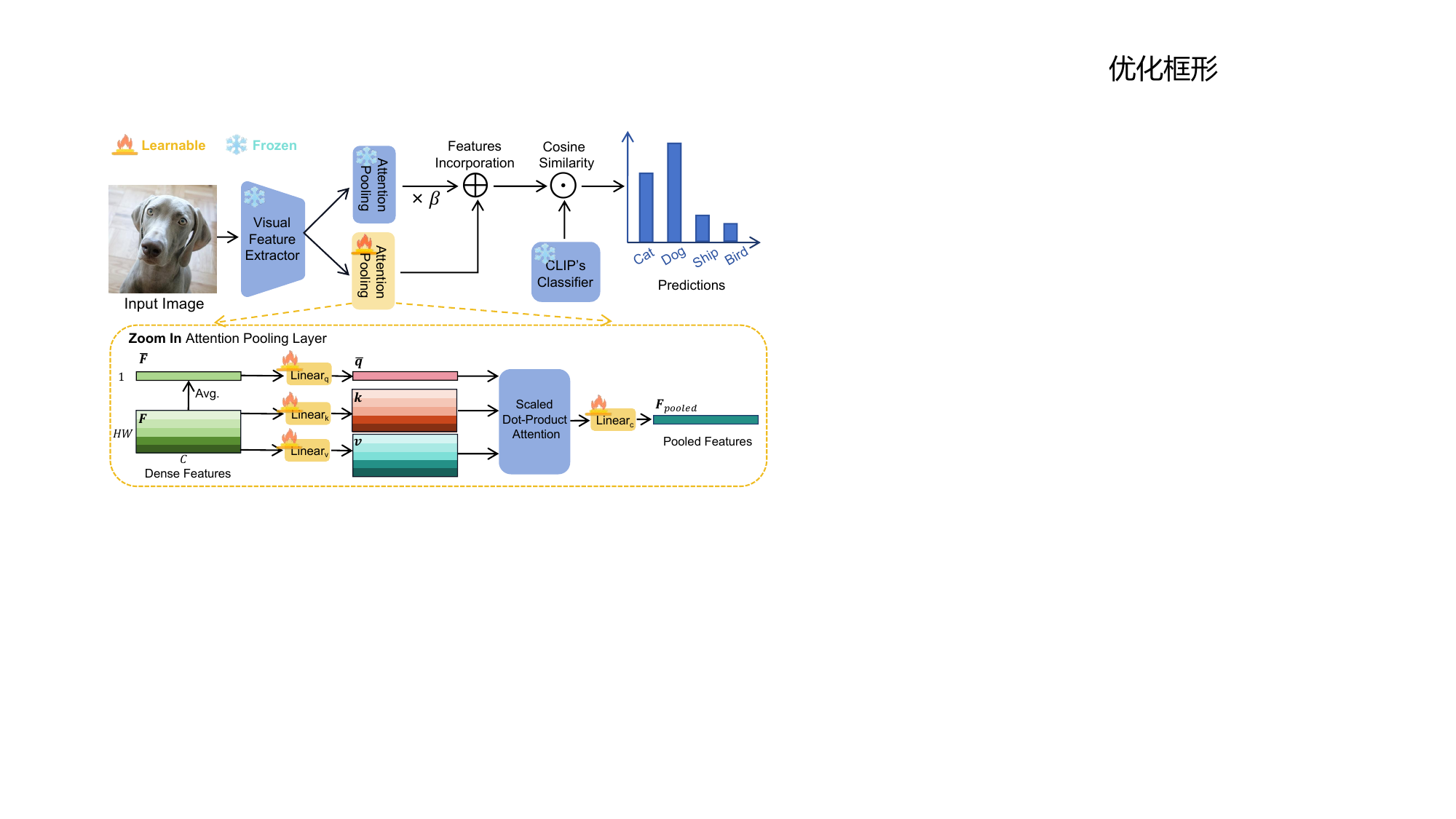}
\centering
\caption{The pipeline of SAFE. During the training process, we only update the parameters of the attention pooling layer of CLIP's visual encoder.
% \wjd{CLIP can take ResNet50 as the backbone. Does ResNet50 have attention pooling layer?}
In the inference process, we perform residual blending between the features pooled by the fine-tuned and the original attention pooling layer. }
\label{Fig:method}
\end{figure*}

\subsection{Semantic-Aware Fine-tuning}\label{SAFE}

This section revisits the CLIP's visual encoder, aiming to exploit the semantic information within dense features to further unleash CLIP's potential in few-shot classification.
As shown in Fig.\ref{Fig:method}, CLIP's visual encoder (with ResNet-50 as its backbone) \cite{radford2021CLIP}\footnote{https://github.com/OpenAI/CLIP} consists of a visual feature extractor and a unique attention pooling layer, which utilizes a transformer-style multi-head attention mechanism. 
The operations within the attention pooling layer are as follows: the global average feature $\overline{\boldsymbol{F}}$ is mapped to the query $\overline{\boldsymbol{q}}$ through the linear layer $\text{Linear}_q$, and dense features ${\boldsymbol{F}}$ at each spatial position are mapped to key-value pairs through the linear layer $\text{Linear}_k$ and $\text{Linear}_v$.
Then, the attention pooling layer obtains the spatial weighted sum of the incoming dense features through the scaled dot-product attention followed by a linear layer $\text{Linear}_c$. This process can be formulated as:
%\wjdd{What is Linear\_c?}
% \begin{equation}
%     \text{AttnPool}(\overline{\boldsymbol{q}},\boldsymbol{k},\boldsymbol{v})=\text{Linear}_c\big(\sum_{j=1}^{HW} \text{softmax} (\frac{\overline{\boldsymbol{q}}\boldsymbol{k}_j^T}{C})\boldsymbol{v}_j \big),
% \end{equation} \label{attention}
% \begin{equation}
%     \overline{\boldsymbol{q}} = \text{Linear}_q(\overline{\boldsymbol{f}}), 
%     \boldsymbol{k}_j = \text{Linear}_k(\boldsymbol{f}_j),
%     \boldsymbol{v}_j = \text{Linear}_v(\boldsymbol{f}_j),
% \end{equation} 

% \begin{spacing}{1.5}

 \begin{equation}
 \resizebox{0.91\hsize}{!}{$
 \left\{
 \begin{array}{l}   
    \boldsymbol{F}_{pooled}=\text{AttnPool}(\boldsymbol{F})=\text{Linear}_c\big(\sum_{j=1}^{HW} \text{softmax} (\frac{\overline{\boldsymbol{q}}\boldsymbol{k}_j^T}{S})\boldsymbol{v}_j \big),\\
    \overline{\boldsymbol{q}} = \text{Linear}_q(\overline{\boldsymbol{F}}), 
    \boldsymbol{k}_j = \text{Linear}_k(\boldsymbol{F}_j),
    \boldsymbol{v}_j = \text{Linear}_v(\boldsymbol{F}_j),  
 \end{array}
 \right.
 $}
 \label{attention}
 \end{equation}
 % \end{spacing}
where $H$ and $W$ represent the height and width of dense features, $S$ is a constant scaling factor, $\boldsymbol{F}_j$ represents the input feature at spatial location $j$ and $\overline{\boldsymbol{F}}$ is the average of all $\boldsymbol{F}_j$. The outputs of the attention pooling layer serve as a comprehensive representation of the whole image, which can capture crucial semantics in dense feature maps for visual recognition.

To put it succinctly, the attention pooling layer assigns different weights to various spatial semantics within dense features, getting pooled features through a weighted-sum approach. The attention pooling layer pre-trained on large-scale datasets focuses on various semantics of images,  providing the model with generality. However, for downstream tasks, task-specific semantics might exert a more pronounced positive influence on classification performance.
Consequently, we propose fine-tuning the parameters in the attention pooling layer while freezing the remaining parameters in CLIP's visual encoder. This process encourages the model to focus on semantics relevant to specific few-shot adaptation tasks.
In consideration of the potential catastrophic forgetting issue \cite{serra2018overcoming,gao2022bmu} caused by fine-tuning, in the inference process, we perform residual blending between the fine-tuned attention pooling layer and the original attention pooling layer to incorporate the prior knowledge from the pre-training and the few-shot knowledge. 

We denote the fine-tuned attention pooling layer as $\text{AttnPool}_\text{F}$, the original attention pooling layer as $\text{AttnPool}_\text{O}$, and the dense features extracted by CLIP's visual feature extractor as $\boldsymbol{F}$. 
We denote the weights of CLIP’s classifier generated from its pre-trained textual encoder as $\boldsymbol{W}_c$. The residual blended feature and output logits can be formulated as follows:
 \begin{equation}\label{Eq:blending_feature}
    \boldsymbol{F}_{blending} = \beta*\text{AttnPool}_\text{O}(\boldsymbol{F})+\text{AttnPool}_\text{F}(\boldsymbol{F}),
\end{equation}
\begin{equation}
    \text{logits} = \boldsymbol{F}_{blending}\boldsymbol{W}_c^T,
\end{equation} 
where $\beta$ denotes the residual ratio. 

Furthermore, the last Transformer layer in the CLIP's visual encoder \cite{radford2021CLIP} with Vision Transformer (ViT) architecture closely resembles the attention pooling layer in the ResNet architecture. The distinctions between them are minimal: first, the global query in ViT is generated by a special class token instead of the average among all spatial locations; second, the Transformer layer incorporates a residual connection. Hence, when using the ViT architecture to implement the CLIP's visual encoder, we treat the last transformer layer the same as the attention pooling layer and fine-tune this layer to enhance the few-shot adaptation performance.
We term our proposed approach as \textbf{S}emantic-\textbf{A}ware \textbf{F}in\textbf{E}-tuning (SAFE) and detail the above procedure in Alg.\ref{algorithm:SAFE}. 

 \begin{algorithm}[ht]
    % \vskip 0.2in    
    \caption{\textbf{S}emantic-\textbf{A}ware \textbf{F}in\textbf{E}-tuning (\textbf{SAFE}).}
    \begin{algorithmic}
    \STATE {$\blacktriangleright$ \textbf{Require}}: 
   The feature extractor $\text{G}$, the frozen original attention pooling layer $\text{AttnPool}_\text{O}$ and the fine-tuned attention pooling layer $\text{AttnPool}_\text{F}$, the weights of CLIP's classifier $\boldsymbol{W}_c$, iterations $N$, residual ratio $\beta$.

  \STATE
  \STATE
    {$\blacktriangleright$ \textbf{Fine-tuning}}:
        \STATE We omit the early stopping process with validation performance for simplicity.
        \FOR {$i = 1, 2, ..., N$}
            \STATE
            $\boldsymbol{x}$, $y$ = few$\_$shot$\_$loader.next()
            \STATE
            $\boldsymbol{F}_{pooled}$ = $\text{AttnPool}_\text{F}$(\text{G}($\boldsymbol{x}$))
            \STATE 
            loss = Cross$\_$Entropy($\boldsymbol{F}_{pooled}W_c^T$, $y$)
            \STATE 
            loss.backward()
            \STATE
            update($\text{AttnPool}_\text{F}$.params)
    	\ENDFOR
    	\STATE \textbf{return} The fine-tuned $\text{AttnPool}_\text{F}$.
    	
  \STATE
  \STATE
    {$\blacktriangleright$ \textbf{Inference Process}}:
        \STATE Given a test sample $\boldsymbol{x}_t$.
        \STATE
        $\boldsymbol{F}$ = G($\boldsymbol{x}_t$)
        \STATE 
        $\boldsymbol{F}_{blending}$ = $\beta*\text{AttnPool}_\text{O}(\boldsymbol{F})+\text{AttnPool}_\text{F}(\boldsymbol{F})$
        \STATE
        logits = $\boldsymbol{F}_{blending}\boldsymbol{W}_c^T$
        \STATE \textbf{return} logits
    \end{algorithmic}
    \label{algorithm:SAFE}
\end{algorithm}

Our approach enhances the few-shot performance of CLIP by refining its attention pooling layer, and this operation is compatible with existing methods that use adapters to enhance CLIP. Taking TIP-Adapter \cite{zhang2022tipadapter} as an example, it constructs an adapter using a key-value cache model derived from pooled features of a few-shot training set. 
Specifically, given K-shot N-class training samples for few-shot classification (denoted as $\boldsymbol{X}_K$ with
their labels $L_N$), TIP-Adapter utilizes the CLIP’s pre-trained visual encoder to extract their C-dimensional features and convert their ground-truth labels into N-dimensional one-hot vectors. These features $\boldsymbol{F}_{Train}\in \mathbb{R}^{NK*C}$ and label vectors $\boldsymbol{L}_{Train}\in \mathbb{R}^{NK*N}$ can be expressed as:
\begin{equation}\label{Eq:F_Train}
 % \left\{
 % \begin{array}{l}  
\boldsymbol{F}_{Train} = \text{AttnPool}_\text{O}(\text{G}(\boldsymbol{X}_K)),
% \end{array}
% \right.
\end{equation}
\begin{equation}
 % \left\{
 % \begin{array}{l}  
\boldsymbol{L}_{Train} = \text{OneHot}(L_N),
% \end{array}
% \right.
\end{equation}
where $\text{G}$ represents the feature extractor.
Then TIP-Adapter treats $\boldsymbol{F}_{Train}$ as keys, while the label vectors $\boldsymbol{L}_{Train}$ as values to construct the cache model, which memorizes all the new knowledge extracted from the few-shot training samples. During the inference process, TIP-Adapter first extracts the C-dimensional feature of the given test sample $\boldsymbol{x}_t$:
\begin{equation}\label{Eq:test_feature}
\boldsymbol{f}_{\boldsymbol{x}_t} = \text{AttnPool}_\text{O}(\text{G}(\boldsymbol{x}_t)).
\end{equation}
TIP-Adapter incorporates the few-shot knowledge retrieved from the cache model and the prior knowledge of pre-trained CLIP to calculate the output logits of $\boldsymbol{x}_t$:
\begin{equation}\label{Eq:tip-f}
\text{logits} = \alpha\phi(\boldsymbol{f}_{\boldsymbol{x}_t}\boldsymbol{F}^T_{Train})\boldsymbol{L}_{Train} + \boldsymbol{f}_{\boldsymbol{x}_t}W_c^T,
\end{equation}
where $\phi(x)=\text{exp}(-\gamma(1-x))$, $\gamma$ and $\alpha$ denote adjustable hyper-parameters. Our proposed SAFE is compatible with this type of adapter approach because these methods enhance few-shot CLIP from different perspectives. Intuitively, compared to the original CLIP model, which calculates logits solely through $\boldsymbol{f}_{\boldsymbol{x}_t}W_c^T$, TIP-Adapter incorporates a cache model to retrieve few-shot knowledge, leading to enhanced performance. Our approach fine-tunes the visual encoder to focus on task-specific semantics, thereby outputting beneficial pooled features for few-shot tasks.
Combining Tip-Adapter with our SAFE approach is straightforward. It only requires replacing the original pooling operation in Eq. (\ref{Eq:F_Train}) and Eq. (\ref{Eq:test_feature}) with our refined pooling operation as Eq. (\ref{Eq:blending_feature}) while keeping the rest of the computation unchanged.
In this way, the keys $\boldsymbol{F}_{Train}$ of the cache model and the features $\boldsymbol{f}_{\boldsymbol{x}_t}$ of the given test sample are reformulated as:
{
\begin{equation}\label{Eq:F_Train_new}
\resizebox{0.91\hsize}{!}{$
\begin{aligned}
\boldsymbol{F}_{Train} = \beta*\text{AttnPool}_\text{O}(\text{G}(\boldsymbol{X}_K))+\text{AttnPool}_\text{F}(\text{G}(\boldsymbol{X}_K)),
\end{aligned}
$}
\end{equation}
}
\begin{equation}\label{Eq:test_feature_new}
\boldsymbol{f}_{\boldsymbol{x}_t} = \beta*\text{AttnPool}_\text{O}(\text{G}(\boldsymbol{x}_t)) +\text{AttnPool}_\text{F}(\text{G}(\boldsymbol{x}_t)).
\end{equation}
Then the output logits of the test sample $\boldsymbol{x}_t$ can still be calculated by Eq. (\ref{Eq:tip-f}).
% \begin{equation}
% \text{logits} = \alpha\phi(\boldsymbol{f}_{\boldsymbol{x}_t}^{SAFE}\boldsymbol{F}_{Train}^{SAFE})\boldsymbol{L}_{Train} + \boldsymbol{f}_{\boldsymbol{x}_t}^{SAFE}W_c^T.
% \end{equation}
This simple modification resulted in a 3.78$\%$ average performance improvement across 11 datasets in the 1-shot scenario than the original TIP-Adapter-F \cite{zhang2022tipadapter}.
We refer to this method as SAFE-A in the following sections.

Up to now, we have presented the exposition on the algorithm's motivation and methodology. Different from the conventional practices of classifier-finetuning \cite{lin2023multimodality,radford2021CLIP}, full-finetuning \cite{wortsman2022robust,radford2021CLIP}, adapter-finetuning \cite{zhang2022tipadapter,gao2023clipadapter} and prompt tuning \cite{pratt2023does,zhou2022learning}, SAFE only fine-tunes the attention pooling layer to concentrate on semantics pertinent to downstream tasks and perform residual blending on the features pooled by the original and fine-tuned attention pooling layer to enhance generalization capability. This approach, marked by its simplicity and elegance, effectively reconciles the conflict arising from the scarcity of data and the magnitude of model parameters and achieves superior performance in few-shot adaptation tasks.

 \section{Experiments} \label{sec:experiments}
 
In Section \ref{sec:setting}, we first present the detailed settings of SAFE and SAFE-A. Then, in Section \ref{sec:fewshot}, we evaluate our approach on 11 widely adopted benchmarks. After that, we evaluate the out-of-distribution performance of our approach by fine-tuning the model in in-domain ImageNet and assessing the generalization capability on out-of-distribution datasets.

\begin{table*}[htbp]
 \centering
\caption{Per-dataset results on the ResNet-50 backbone. We bold the best result for each shot and each dataset. SAFE-A consistently produces the best performance across all datasets.}
\scalebox{0.8}{ 
\begin{tabular}{c|c|cccccccccccc}
\hline
\multirow{2}{*}{\textbf{Method}} & \multirow{2}{*}{\textbf{shots}} & \multicolumn{11}{c}{\textbf{Datasets}} & \multicolumn{1}{l}{} \\ \cline{3-14} 
 &  & Aircraft & ImageNet & DTD & EuroSAT & Caltech & Food & Flowers & Pets & Cars & SUN397 & \multicolumn{1}{c|}{UCF101} & \multicolumn{1}{l}{Average} \\ \hline
Zero-Shot CLIP \cite{radford2021CLIP}& 0 & 17.28 & 58.18 & 42.32 & 37.56 & 86.29 & 77.31 & 66.14 & 85.77 & 55.61 & 58.52 & \multicolumn{1}{c|}{61.46} & 58.77 \\ \hline
CuPL \cite{pratt2023does}& 0 & 19.53 & 61.67 & 48.46 & 37.99 & 89.17 & 77.07 & 65.97 & 85.04 & 57.28 & 62.75 & \multicolumn{1}{c|}{59.71} & 60.42 \\ \hline
\multirow{5}{*}{WiSE-FT \cite{wortsman2022robust}} & 1 & 18.61 & 58.30 & 44.17 & 52.30 & 85.49 & 71.88 & 65.83 & 81.73 & 55.66 & 56.59 & \multicolumn{1}{c|}{59.39} & 59.09 \\
 & 2 & 20.88 & 59.08 & 46.95 & 57.07 & 87.00 & 73.54 & 71.02 & 82.75 & 58.67 & 60.15 & \multicolumn{1}{c|}{62.74} & 61.80 \\
 & 4 & 23.33 & 60.48 & 52.23 & 62.45 & 89.03 & 76.17 & 77.10 & 85.95 & 62.09 & 63.18 & \multicolumn{1}{c|}{66.14} & 65.29 \\
 & 8 & 26.97 & 61.85 & 55.56 & 71.40 & 90.07 & 76.72 & 82.54 & 86.52 & 66.00 & 65.25 & \multicolumn{1}{c|}{69.84} & 68.43 \\
 & 16 & 31.75 & 62.84 & 61.74 & 77.79 & 90.79 & 77.80 & 86.91 & 87.50 & 71.28 & 67.46 & \multicolumn{1}{c|}{72.20} & 71.64 \\ \hline
\multirow{5}{*}{CoOp \cite{zhou2022learning}} & 1 & 9.64 & 57.15 & 44.39 & 50.63 & 87.53 & 74.32 & 68.12 & 85.89 & 55.59 & 60.29 & \multicolumn{1}{c|}{61.92} & 59.59 \\
 & 2 & 18.68 & 57.81 & 45.15 & 61.50 & 87.93 & 72.49 & 77.51 & 82.64 & 58.28 & 59.48 & \multicolumn{1}{c|}{64.09} & 62.32 \\
 & 4 & 21.87 & 59.99 & 53.49 & 70.18 & 89.55 & 73.33 & 86.20 & 86.70 & 62.62 & 63.47 & \multicolumn{1}{c|}{67.03} & 66.77 \\
 & 8 & 26.13 & 61.56 & 59.97 & 76.73 & 90.21 & 71.82 & 91.18 & 85.32 & 68.43 & 65.52 & \multicolumn{1}{c|}{71.94} & 69.89 \\
 & 16 & 31.26 & 62.95 & 63.58 & 83.53 & 91.83 & 74.67 & 94.51 & 87.01 & 73.36 & 69.26 & \multicolumn{1}{c|}{75.71} & 73.42 \\ \hline
\multirow{5}{*}{ProGrad \cite{zhu2023prompt}} & 1 & 18.81 & 57.75 & 46.14 & 56.32 & 88.68 & 76.04 & 73.18 & 88.36 & 58.38 & 60.54 & \multicolumn{1}{c|}{64.55} & 62.61 \\
 & 2 & 20.47 & 59.75 & 49.78 & 63.10 & 87.98 & 74.95 & 79.77 & 86.89 & 61.81 & 63.06 & \multicolumn{1}{c|}{66.35} & 64.90 \\
 & 4 & 23.32 & 61.46 & 54.43 & 72.53 & 89.99 & 75.95 & 85.37 & 88.04 & 65.62 & 66.39 & \multicolumn{1}{c|}{69.86} & 68.45 \\
 & 8 & 27.02 & 62.54 & 60.69 & 78.04 & 90.83 & 76.65 & 91.64 & 87.91 & 69.29 & 67.62 & \multicolumn{1}{c|}{73.33} & 71.41 \\
 & 16 & 30.25 & 63.54 & 63.87 & 83.29 & 92.10 & 78.41 & 94.37 & 89.00 & 73.46 & 69.84 & \multicolumn{1}{c|}{75.38} & 73.96 \\ \hline
\multirow{5}{*}{Tip-Adapter-F \cite{zhang2022tipadapter}} & 1 & 20.06 & 60.87 & 48.58 & 51.81 & 87.90 & 77.27 & 76.70 & 86.44 & 58.42 & 62.40 & \multicolumn{1}{c|}{65.38} & 63.26 \\
 & 2 & 21.17 & 61.54 & 51.64 & 66.32 & 89.40 & 77.44 & 79.50 & 86.44 & 61.06 & 63.22 & \multicolumn{1}{c|}{67.45} & 65.93 \\
 & 4 & 24.97 & 62.48 & 57.21 & 69.23 & 90.78 & 77.20 & 89.00 & 86.45 & 64.54 & 65.75 & \multicolumn{1}{c|}{71.17} & 68.98 \\
 & 8 & 28.13 & 63.94 & 61.92 & 77.69 & 91.10 & 78.36 & 92.40 & 88.11 & 69.32 & 68.28 & \multicolumn{1}{c|}{74.42} & 72.15 \\
 & 16 & 34.83 & 65.18 & 66.23 & 81.96 & 92.28 & 79.05 & 93.90 & 89.13 & 75.08 & 71.27 & \multicolumn{1}{c|}{77.24} & 75.10 \\ \hline
\multirow{5}{*}{Cross-Modal \cite{lin2023multimodality}} & 1 & 20.56 & 61.55 & 49.92 & 61.84 & 89.10 & 77.14 & 76.25 & 85.72 & 58.96 & 63.38 & \multicolumn{1}{c|}{66.80} & 64.66 \\
 & 2 & 22.65 & 62.64 & 55.18 & 68.48 & 89.97 & 78.19 & 82.80 & 87.24 & 61.19 & 65.81 & \multicolumn{1}{c|}{70.34} & 67.68 \\
 & 4 & 25.58 & 62.77 & 60.68 & 75.21 & 91.30 & 78.57 & 88.66 & 87.86 & 64.49 & 67.76 & \multicolumn{1}{c|}{73.61} & 70.59 \\
 & 8 & 33.87 & 64.23 & 64.72 & 81.33 & 92.20 & 78.92 & 93.50 & 88.71 & 69.06 & 69.64 & \multicolumn{1}{c|}{77.50} & 73.97 \\
 & 16 & 43.60 & 65.95 & 68.91 & 86.67 & 93.52 & 78.66 & 95.72 & 89.12 & 75.45 & 71.91 & \multicolumn{1}{c|}{79.95} & 77.22 \\ \hline
\multirow{5}{*}{\textbf{\begin{tabular}[c]{@{}c@{}}SAFE\\ (Ours)\end{tabular}}} & 1 & 22.67 & 61.85 & 54.26 & 64.93 & 90.07 & 77.50 & 77.57 & 87.37 & 59.66 & 64.74 & \multicolumn{1}{c|}{67.30} & 66.17 \\
 & 2 & 24.76 & 62.64 & 56.61 & 69.96 & 90.69 & 77.99 & 82.86 & 87.81 & 61.56 & 66.68 & \multicolumn{1}{c|}{71.16} & 68.43 \\
 & 4 & 27.27 & 63.03 & 62.47 & 75.24 & 91.72 & 78.59 & 88.64 & 88.52 & 63.87 & 67.82 & \multicolumn{1}{c|}{74.61} & 71.07 \\
 & 8 & 33.81 & 64.86 & 65.96 & 81.77 & 92.79 & 78.78 & 93.49 & 89.33 & 67.30 & 69.58 & \multicolumn{1}{c|}{78.06} & 74.16 \\
 & 16 & 42.03 & 65.93 & 69.94 & 86.97 & 93.94 & 79.66 & 95.82 & 89.67 & 73.23 & 72.17 & \multicolumn{1}{c|}{81.24} & 77.33 \\ \hline
\multirow{5}{*}{\textbf{\begin{tabular}[c]{@{}c@{}}SAFE-A\\ (Ours)\end{tabular}}} & 1 & \textbf{23.82} & \textbf{61.88} & \textbf{56.03} & \textbf{65.30} & \textbf{90.07} & \textbf{77.69} & \textbf{80.55} & \textbf{88.06} & \textbf{61.52} & \textbf{64.76} & \multicolumn{1}{c|}{\textbf{67.73}} & \textbf{67.04} \\
 & 2 & \textbf{26.01} & \textbf{62.83} & \textbf{57.45} & \textbf{71.27} & \textbf{90.69} & \textbf{78.03} & \textbf{84.61} & \textbf{88.28} & \textbf{64.88} & \textbf{67.23} & \multicolumn{1}{c|}{\textbf{71.23}} & \textbf{69.32} \\
 & 4 & \textbf{28.95} & \textbf{63.52} & \textbf{62.94} & \textbf{76.59} & \textbf{92.01} & \textbf{78.72} & \textbf{89.04} & \textbf{88.74} & \textbf{67.96} & \textbf{68.52} & \multicolumn{1}{c|}{\textbf{74.62}} & \textbf{71.96} \\
 & 8 & \textbf{34.02} & \textbf{65.11} & \textbf{66.08} & \textbf{83.44} & \textbf{92.79} & \textbf{79.17} & \textbf{93.99} & \textbf{89.42} & \textbf{70.31} & \textbf{70.79} & \multicolumn{1}{c|}{\textbf{78.56}} & \textbf{74.88} \\
 & 16 & \textbf{44.10} & \textbf{66.52} & \textbf{70.63} & \textbf{87.50} & \textbf{93.94} & \textbf{79.75} & \textbf{95.93} & \textbf{89.74} & \textbf{77.03} & \textbf{72.72} & \multicolumn{1}{c|}{\textbf{81.50}} & \textbf{78.12} \\ \hline
\end{tabular}
} \label{tab:rn50_fewshot}
\end{table*}

\begin{table}[htbp]
 \centering
\caption{Detailed statistics of these 11 datasets.}
\scalebox{0.92}{ 
\begin{tabular}{c|cccc}
\hline
Dataset       & Classes & Train & Val   & Test  \\
\hline
ImageNet \cite{deng2009imagenet} & 1000    & 1.28M & N/A   & 50000 \\
DTD \cite{cimpoi2014describing} & 47      & 2820  & 1128  & 1692  \\
Stanford-Cars \cite{krause20133d}& 196     & 6509  & 1635  & 8041  \\
Oxford-Pets \cite{parkhi2012cats}& 37      & 2944  & 736   & 3669  \\
UCF101 \cite{soomro2012ucf101}& 101     & 7639  & 1898  & 3783  \\
Caltech101 \cite{fei2004learning}& 100     & 4128  & 1649  & 2465  \\
Flowers102 \cite{nilsback2008automated}& 102     & 4093  & 1633  & 2463  \\
Food101 \cite{bossard2014food}& 101     & 50500 & 20200 & 30300 \\
SUN397 \cite{xiao2010sun}& 397     & 15880 & 3970  & 19850 \\
EuroSAT \cite{helber2019eurosat}& 10      & 13500 & 5400  & 8100  \\
FGCV-Aircraft \cite{maji2013fine}& 100     & 3334  & 3333  & 3333 \\
\hline
\end{tabular}
} \label{tab:datasets}
\end{table}

\subsection{Settings}\label{sec:setting}
\textbf{Datasets.} We conduct few-shot experiments on 11 widely adopted datasets: ImageNet \cite{deng2009imagenet}, DTD \cite{cimpoi2014describing}, Standford-Cars \cite{krause20133d}, Oxford-Pets \cite{parkhi2012cats}, UCF101 \cite{soomro2012ucf101}, Caltech101 \cite{fei2004learning}, Flowers102 \cite{nilsback2008automated}, Food101 \cite{bossard2014food}, SUN397 \cite{xiao2010sun}, EuroSAT \cite{helber2019eurosat}, and FGVC-Aircraft \cite{maji2013fine}. To ensure reproducibility, we use CoOp’s \cite{zhou2022learning} dataset splits and the three-fold few-shot train sets sampled with the same random seeds to fine-tune the CLIP with 1, 2, 4, 8, and 16 labeled samples per class. We detail the statics of these 11 datasets in Tab. \ref{tab:datasets}.

\textbf{Few-shot evaluation protocol.} To ensure a fair comparison, we strictly follow the protocol of CoOp \cite{zhou2022learning}, and Cross-Modal Fine-tuning \cite{lin2023multimodality} by reporting test performance on 11 public image datasets, with ResNet-50 as the visual encoder backbone of CLIP by default. We have also evaluated the effectiveness of our method with ViT-B/16 as the visual encoder. We adopt the full test set of the downstream task to evaluate each method. We augment the class names into sentences using hand-engineered templates suggested by Tip-Adapter \cite{zhang2022tipadapter} to get the textual prompt for initializing CLIP's classifier. We discuss the influence of different prompts in Section \ref{sec:prompt}.  

\textbf{Training details.} 
Following existing practices \cite{zhang2022tipadapter,lin2023multimodality}, we use the AdamW optimizer and implement cosine annealing learning rate scheduling over the course of 12,800 iterations for all experiments. The batch size is set to 8. We conduct a grid search for the learning rate in the range of [0.0001, 0.00001, 0.000001, 0.0000001] and weight decay in the range of [0.0, 0.001, 0.00001] for each task. Early stopping is applied based on the evaluation performance every 100 iterations on the given few-shot validation set with a minimum of ($n$, 4) samples, where $n$ represents the number of training shots. All experiments are conducted on a single Tesla V100 GPU. The hyperparameter residual ratio $\beta$ is set to 0.5 by default. We discuss the influence of the hyperparameter in Section \ref{sec:hyper}.

\subsection{Few-shot Performance}\label{sec:fewshot}

As shown in Tab. \ref{tab:rn50_fewshot}, both SAFE and SAFE-A exhibit exceptional performance compared to other methods. The zero-shot CLIP numbers here slightly differ from those reported in the original CLIP paper \cite{radford2021CLIP} because we use the hand-engineered prompt adopted in existing few-shot adaptation works \cite{zhang2022tipadapter,lin2023multimodality} rather than the original prompt. The reported Cross-Modal \cite{lin2023multimodality} results here are based on the best few-shot performance settings suggested in their paper. For the average performance across 11 few-shot datasets, SAFE outperforms the CoOp method by +6.58$\%$ and the Cross-Modal by +1.51$\%$ in the 1-shot setting. Notably, even with just 1 shot, SAFE outperforms the 16-shot CoOp methods by +2.83$\%$ on the Food101 dataset and +0.36$\%$ on the Oxford-Pets dataset.

In some scenarios, SAFE's performance may be inferior to existing methods. For instance, on the Stanford-Cars dataset, SAFE's performance is 1.85$\%$ lower than that of TIP-Adapter-F in the 16-shot setting, while outperforming TIP-Adapter-F by +1.24$\%$ in the 1-shot setting. We think this is because as the number of training samples increases, the cache in the TIP-Adapter stores sufficient few-shot knowledge to enhance the classification task. Luckily, SAFE is an elegant plug-and-play method, which can be combined with existing approaches to further boost the few-shot adaptation performance. With the assistance of adapter methods, SAFE-A achieves the best performance across all few-shot settings when compared to existing methods. On the Stanford-Cars dataset, SAFE-A outperforms TIP-Adapter-F by +3.10$\%$ in the 1-shot setting and +1.95$\%$ in the 16-shot setting. Notably, even with just 1-shot, SAFE-A outperforms the 16-shot CoOp and WiSE-FT methods by +1.05$\%$ and +0.56$\%$, respectively, on the Oxford-Pets dataset. 
These superior results fully demonstrate the effectiveness of encouraging the model to focus on task-specific semantics in few-shot adaptation tasks.

\subsection{Out-of-distribution Performance}

In addition to the performance on the specific downstream task (speciality), the model's generality on out-of-distribution datasets is also a vital issue. The CLIP model is pre-trained on vast datasets, which endows it with good generality. In this section, we evaluate whether our proposed method can enhance CLIP's out-of-distribution performance while improving its few-shot performance. We follow the protocol of existing works \cite{zhou2022learning, lin2023multimodality} for evaluating the out-of-distribution performance. We set ImageNet \cite{deng2009imagenet} as the source dataset with 16-shot training samples, and test the performance against test-time distribution shifts on variants of ImageNet: ImageNetV2 \cite{recht2019imagenet}, ImageNet-A \cite{hendrycks2021natural}, ImageNet-R \cite{hendrycks2021many} and ImageNet-Sketch \cite{wang2019learning}. The reported Cross-Modal \cite{lin2023multimodality} results here are based on the best out-of-distribution performance settings suggested in their paper.

\begin{table}[htbp]
 \centering
\caption{Comparison of domain generalization performance ($\%$) of SAFE and SAFE-A. We utilize 16-shot ImageNet as the training data before the out-of-distribution test.}
\scalebox{0.80}{ 
\begin{tabular}{cccccc}
\hline
\multirow{2}{*}{Method} & Source & \multicolumn{4}{c}{Target} \\ \cline{2-6} 
 & ImageNet & -V2 & -Sketch & -A & -R \\ \hline
\multicolumn{6}{l}{\textbf{ResNet-50}} \\
Zero-Shot CLIP \cite{radford2021CLIP} & 58.2 & 51.3 & 33.3 & 21.7 & 56.0 \\
Linear Probing \cite{radford2021CLIP}& 55.9 & 46.0 & 19.1 & 12.7 & 34.9 \\
CoOp(M=4) \cite{zhou2022learning}& 63.0 & 55.1 & 32.7 & 22.1 & 55.0 \\
CoOp(M=16) \cite{zhou2022learning}& 63.3 & 55.4 & 34.7 & 23.1 & 56.6 \\
WiSE-FT \cite{wortsman2022robust}& 62.9 & 54.2 & 33.3 & 20.3 & 57.4 \\
TIP-Adapter-F \cite{zhang2022tipadapter}& 65.2  & 57.1&36.0&22.8&58.5\\
Cross-Modal \cite{lin2023multimodality}& 64.5 & 55.3 & 33.1 & 20.0 & 56.4 \\
Cross-Modal+WiSE-FT \cite{lin2023multimodality}& 65.2 & 56.6 & 35.6 & 22.6 & 59.5 \\
CALIP-FS \cite{guo2023calip}& 65.8 & 55.9 & 35.4 & 23.4 & 56.7 \\
\textbf{SAFE(ours)} & 65.9 & 57.7 & \textbf{36.9} & \textbf{23.6} & \textbf{60.3} \\ 
\textbf{SAFE-A(ours)}& \textbf{66.5}& \textbf{58.6}& 36.3& 23.0 & 59.9 
\\ \hline
\multicolumn{6}{l}{\textbf{ViT-B/16}} \\
Zero-Shot CLIP \cite{radford2021CLIP}& 66.7 & 60.8 & 46.2 & 47.8 & 74.0 \\
Linear Probing \cite{radford2021CLIP}& 65.9 & 56.3 & 34.8 & 35.7 & 58.4 \\
CoOp(M=4) \cite{zhou2022learning}& 71.9 & 64.2 & 46.7 & 48.4 & 74.3 \\
CoOp(M=16) \cite{zhou2022learning}& 71.7 & 64.6 & 47.9 & 49.9 & 75.1 \\
CoCoOp \cite{zhou2022conditional}& 71.0 & 64.1 & 48.8 & 50.6 & 76.2 \\
WiSE-FT \cite{wortsman2022robust}& 73.0 & 65.2 & 49.1 & 49.8 & 77.6 \\
Cross-Modal \cite{lin2023multimodality}& 73.2 & 64.8 & 47.9 & 48.3 & 76.7 \\
Cross-Modal+WiSE-FT \cite{lin2023multimodality}& 72.9 & 65.4 & 49.2 & 50.5 & 77.8 \\
CALIP-FS \cite{guo2023calip}& 73.1 & 65.1 & 48.3 & 50.4 & 76.6 \\
\textbf{SAFE(ours)} & 74.2 & 67.0 & \textbf{50.2} & \textbf{51.6} & \textbf{78.1} \\
\textbf{SAFE-A(ours)} & \textbf{75.2} &\textbf{67.8} & 49.8 & 51.3 & 77.4
\\
\hline
\end{tabular} 
} \label{tab:generalization}
\end{table}

We report results with two kinds of visual encoders (ResNet-50 and ViT-B/16) in Tab. \ref{tab:generalization}, and mark the best results in bold. 
Our proposed SAFE method outperforms existing works in terms of out-of-distribution performance across all four test datasets. Specifically, using the ResNet-50 visual encoder, SAFE surpasses the method TIP-Adapter-F by +0.6$\%$ on ImageNet-V2 and outperforms the method Cross-Modal+WiSE-FT by +0.8$\%$ on ImageNet-R. With the VIT-B/16 visual encoder, SAFE exceeds the method Cross-Modal+WiSE-FT by +1.6$\%$ on ImageNet-V2 and outperforms the method CoCoOp by +1.0$\%$ on ImageNet-A.

With the best in-domain performance, SAFE-A achieves significant out-of-distribution performance on ImageNet-V2, which surpasses SAFE by +0.9$\%$ and surpasses the method TIP-Adapter-F by +1.5$\%$ with the ResNet-50 visual encoder. For datasets with more distribution shifts as ImageNet-A, ImageNet-R, and ImageNet-Sketch, SAFE-A performs inferior to SAFE.
Specifically, when using the VIT-B/16 visual encoder, SAFE-A's performance lags behind SAFE by 0.3$\%$ on ImageNet-A and by 0.7$\%$ on ImageNet-R.
We attribute this phenomenon to the adapters utilized in SAFE-A, which is constructed from few-shot training samples. Compared to SAFE, SAFE-A leverages more few-shot knowledge, effectively enhancing the model's performance in scenarios with a similar distribution to the in-domain dataset. However, this approach potentially reduces the model's generality. Consequently, its performance slightly falls behind that of SAFE in scenarios with more distribution shifts.

\subsection{Ablation Study}

\subsubsection{The Influence of the Hyperparamer}\label{sec:hyper}

\begin{figure}[tbh!]
\includegraphics[width=0.43\textwidth]{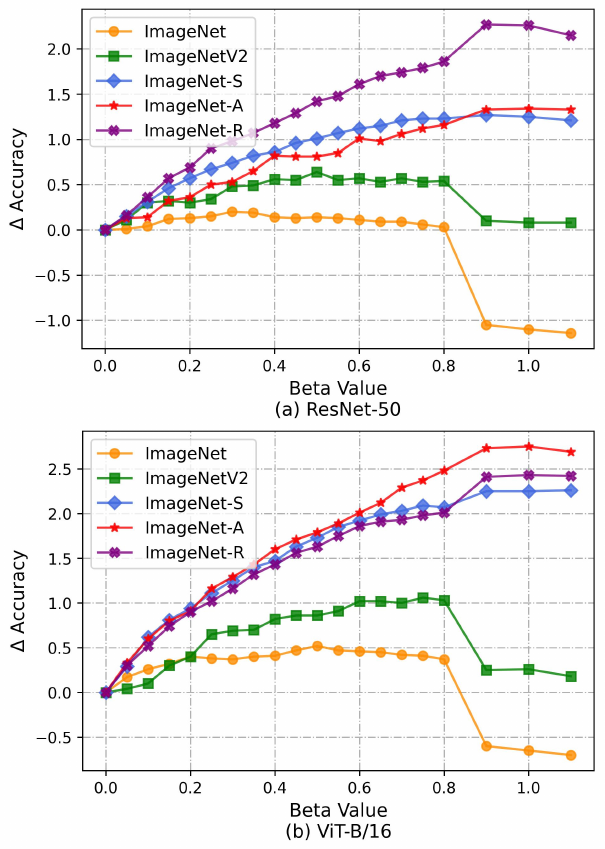}
\centering
\caption{The influence of the hyper-parameter $\beta$. We set ImageNet as the training set with 16-shot and tested the performance on variants of ImageNet. We chose both the ResNet-50 and ViT-B/16 as visual encoder backbones of CLIP.}
\label{Fig:beta}
\end{figure}

Our method introduces only one hyperparameter $\beta$, as shown in Eq. (\ref{Eq:blending_feature}). This subsection discusses the impact of the hyperparameter on few-shot performance, as illustrated in Fig.\ref{Fig:beta}. We fine-tune the CLIP using the 16-shot ImageNet training data and test it on variants of ImageNet. We considered both ResNet-50 and ViT-B/16 as options for the visual backbone. Delta accuracy refers to the performance gain brought by different $\beta$ values, indicating the difference in accuracy compared to the case where $\beta$ is zero. Overall, with the increase in $\beta$, the model's performance shows an upward trend followed by a decline over different datasets. Using ResNet-50 as the backbone, a $\beta$ value of 0.9 yields gains of +2.27$\%$ on the ImageNet-R dataset and +1.27$\%$ on ImageNet-Sketch. As $\beta$ increases further, the gains gradually decrease. On the ImageNet dataset and ImageNet-V2 dataset, performance significantly decreases when $\beta$ exceeds 0.8. Especially for the ImageNet dataset, the performance gain becomes negative when $\beta$ is greater than 0.8. However, for other datasets, the performance decline is not significant and might even continue to improve when $\beta$ exceeds 0.8. For simplicity, we set the hyperparameter to 0.5 in our experiments and have achieved superior performance to existing methods. Further improvement in model performance may be possible with meticulous selection of the hyperparameter $\beta$.

\subsubsection{Can Early Layers of the Visual Encoder Capture Semantics}

Fig. \ref{Fig:motivation} demonstrates that CLIP's feature extractor can capture semantics. Here, the feature extractor refers to the portion before the pooling layer in the visual encoder. Our experiments indicate that fine-tuning the attention pooling layer to focus on task-specific semantics can improve the performance of few-shot CLIP. This leads to a question: \textit{can early layers also capture the semantics, and how about fine-tuning early layers in the visual encoder?}

\begin{figure}[tbh!]
\includegraphics[width=0.46\textwidth]{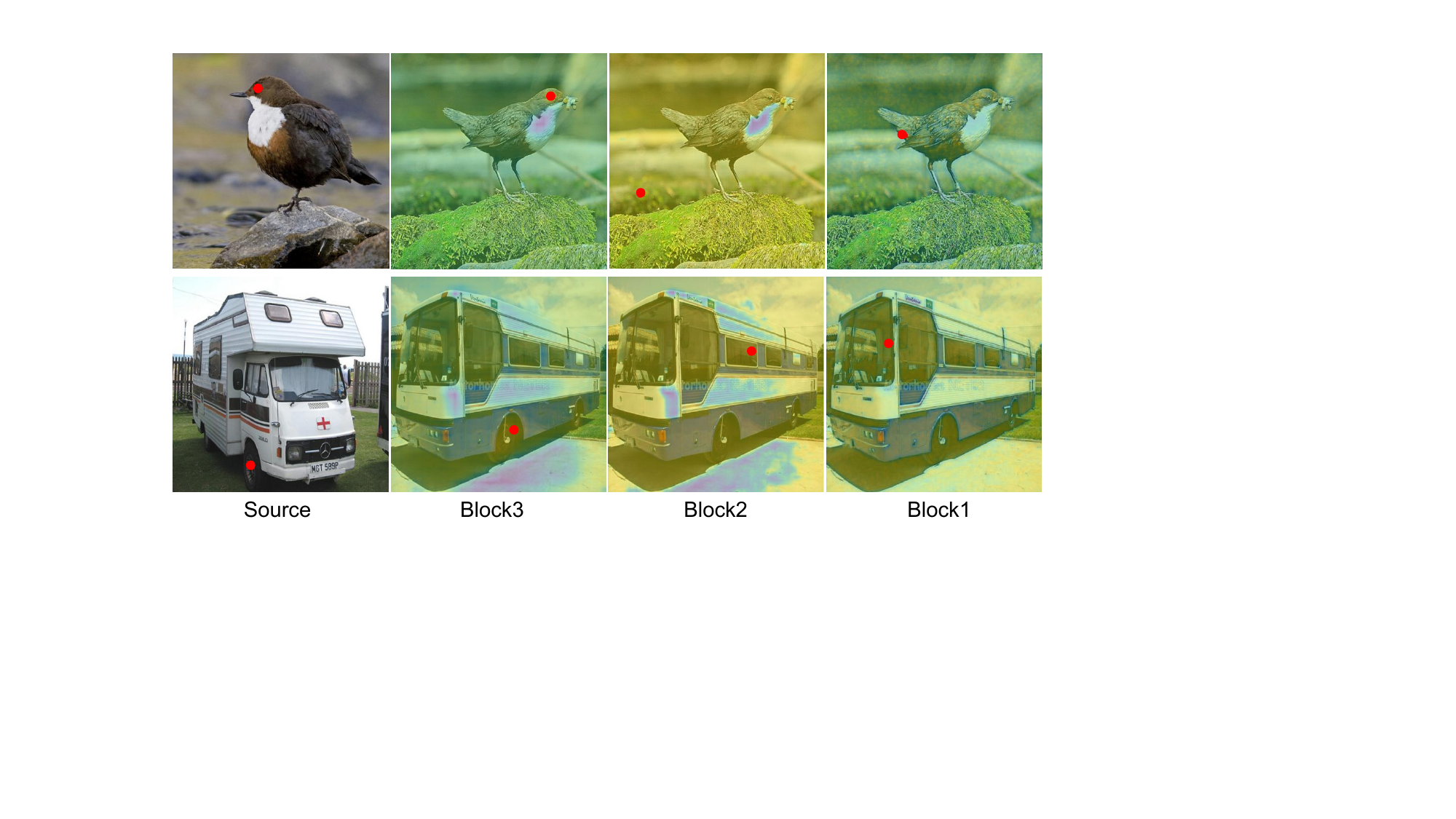}
\centering
\caption{The predicted target point given by different blocks of the visual encoder with ResNet-50 architecture. ``Block3" indicates that we use the output features given by the visual encoder's third residual block to predict the target point.}
\label{Fig:other_layer}
\end{figure}

We chose the ResNet-50 architecture for the CLIP's visual encoder, which consists of four residual blocks and an attention pooling layer. Fig. \ref{Fig:other_layer} qualitatively evaluates the semantic capturing capability of the visual encoder's early layers. The third residual block of the visual encoder demonstrates semantic capturing capabilities; for example, it can correctly predict correspondences between bird eyes. However, the earlier layers show relatively weaker semantic capturing capability. For example, the first residual block incorrectly perceives a strong correlation between a bird's tail and its eyes. Such findings are also consistent with the views of some existing research \cite{natureindividual,zeiler2014visualizing}, which indicates that the earlier layers of the model focus on general features like edges and corners rather than semantic features.

Then, we explore the influence of fine-tuning the early layers. Considering that the third residual block also possesses semantic capturing capabilities, we freeze the first three residual blocks in the visual encoder while simultaneously fine-tuning the fourth residual block and the attention pooling layer with few-shot training samples. As expected, severe overfitting occurred. In such cases, the fine-tuned CLIP's classification performance and out-of-distribution generalization performance sharply decreased compared to the zero-shot CLIP. We attribute this to the fact that the number of parameters involved in fine-tuning far exceeds the training samples. This result indicates that fine-tuning the earlier layers of the CLIP's visual encoder may not be a proper choice for few-shot tasks.

\subsubsection{Class Activation Mapping}

\begin{figure}[tbh!]
\includegraphics[width=0.44\textwidth]{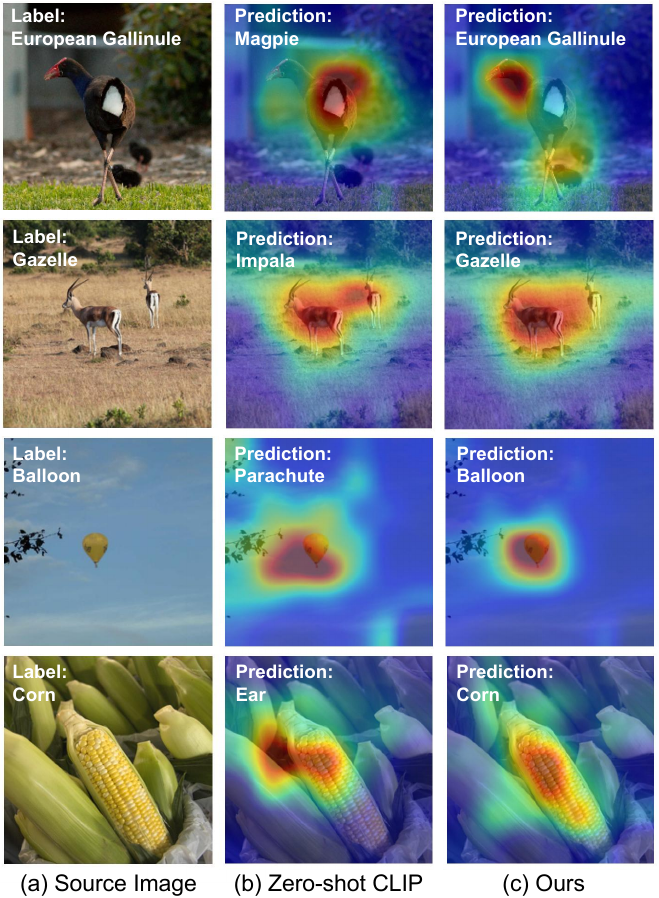}
\centering
\caption{We draw the heat maps to explain which parts of the image dominate the model's decision through Smooth Grad-CAM++\cite{omeiza2019smooth}. The heat map of the CLIP's visual encoder fine-tuned by SAFE for each image aligns better with the objects in the image than that of the original CLIP.}
\label{Fig:cam}
\end{figure}

In this subsection, we use Smooth Grad-CAM++\cite{omeiza2019smooth} to generate heat maps for different images. As shown in Fig. \ref{Fig:cam}, the heat map of our SAFE fine-tuned CLIP's visual encoder aligns better with the objects in the image than that of the original CLIP. We use ResNet-50 as the visual backbone. The heat map of our fine-tuned model for the European gallinule figure shows that the head of this bird dominates the decision, while the original model pays more attention to the tail.
The heat map for the balloon indicates that the original model focuses on an area near the balloon, incorrectly identifying the balloon as a parachute. In contrast, our fine-tuned model directs its attention to the balloon itself, resulting in a correct classification. The model fine-tuned by our method considers more object-relevant parts, which may contribute to its better test accuracy.

\subsubsection{The Influence of the Prompt}\label{sec:prompt}

\begin{figure}[tbh!]
\includegraphics[width=0.48\textwidth]{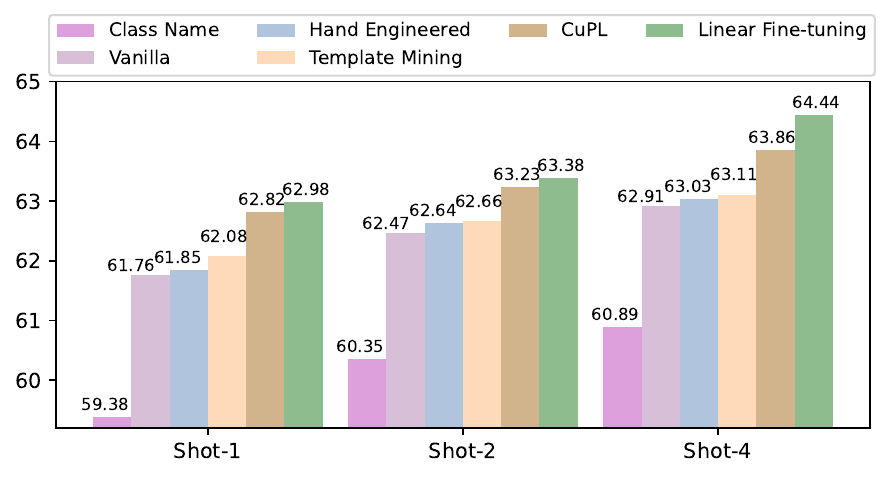}
\centering
\caption{The influence of different prompt choices on our proposed SAFE method. Better prompt choice may lead to further performance improvements.}
\label{Fig:prompt}
\end{figure}

This paper focuses on the impact of the visual encoder on few-shot tasks. We detail the few-shot evaluation protocol in Section \ref{sec:setting}.
% We primarily follow the settings from existing studies \cite{zhang2022tipadapter,lin2023multimodality}, choosing a hand-engineered prompt to generate textual features for initializing CLIP's classifier. 
Prompt designing is also a critical aspect of few-shot tasks. This subsection explores the influence of different prompt choices on few-shot performance. Taking the few-shot task on ImageNet as an example, we illustrate the impact of various prompts on our proposed SAFE method in Fig.\ref{Fig:prompt}. ``Class Name" represents constructing the prompt using only class names. ``Vanilla" indicates using a single prompt, such as ``a photo of a [CLASS]." ``Hand Engineered" indicates adopting prompts suggested by Tip-Adapter \cite{zhang2022tipadapter}. ``Template Mining" involves searching among a pool of 180 templates for 21 templates with the best zero-shot performance, as proposed by \citet{lin2023multimodality}. ``CuPL" utilizes GPT3 to generate specific prompts for each category. ``Linear Fine-tuning" uses textual features from CuPL's prompt to initialize CLIP's classifier and then fine-tunes this linear layer with few-shot samples.

Using only category names as prompts to derive features for initializing CLIP's classifier is not prudent. Across different few-shot settings, its performance lags far behind other methods. The recently proposed CuPL prompts outperform the hand-engineered prompts by +0.97$\%$ in the 1-shot setting and by +0.59$\%$ in the 4-shot setting. Furthermore, in the 1-shot setting, utilizing the linear layer obtained through Linear Fine-tuning as CLIP's classifier, our SAFE's performance increases from 61.85$\%$ to 62.98$\%$ compared to the hand-engineered prompt, resulting in a gain of +1.13$\%$. This highlights the potential of achieving even better few-shot performance through prompt designing. Considering that this paper focuses on revealing the importance of semantics in few-shot tasks and has proposed methods surpassing existing approaches, the question of how to design better prompts is left for future research.

\subsection{Further Discussion}
In this work, we point out that not all pre-trained parameters in the CLIP model are suitable for downstream few-shot tasks. Therefore, freezing its parameters might not be as effective as fine-tuning specific layers to encourage the model to focus on task-specific semantics. Our approach demonstrates satisfactory performance in both few-shot classification and generalization capabilities. Lastly, we share an interesting finding that might draw readers' attention—whether the discoveries in this paper hold true for the recently popular large-scale multimodal models \cite{zhu2023minigpt,liu2023visual}, where the frozen CLIP model could be used as a visual encoder. Using LLaVA-1.5 \cite{liu2023visual} as an example, this model connects CLIP's visual encoder with the language decoder Vicuna \cite{zheng2023judging} through the projection layer. According to our analysis, fine-tuning the last layer of the visual encoder (SAFE method) rather than directly freezing them (Vanilla method) could potentially achieve better performance.

\begin{figure}[tbh!]
\includegraphics[width=0.45\textwidth]{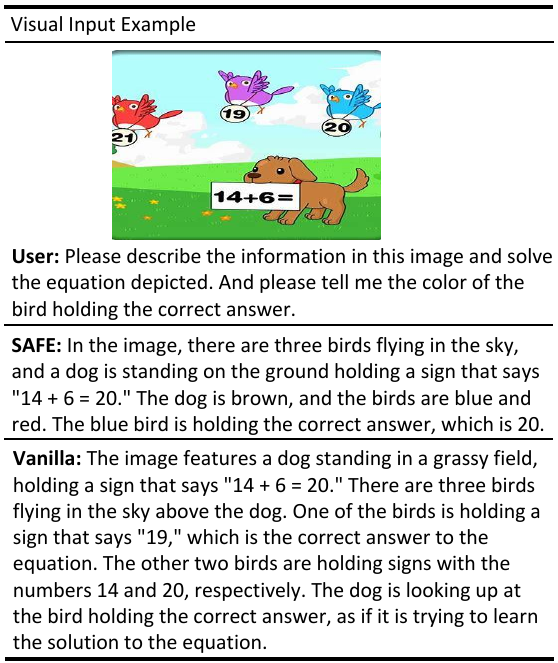}
\centering
\caption{Example prompt demonstrates the visual input capability of the vanilla LLaVA and the LLaVA trained through the proposed SAFE method.}
\label{Fig:llava}
\end{figure}

Qualitatively, using the method proposed in this paper to train the LLaVA model can somewhat enhance its capability in handling visual-textual dialogue tasks, for example, yielding more reasonable responses in Fig. \ref{Fig:llava}.

\begin{table}[htbp]
 \centering
\caption{Comparison between the Vanilla LLaVA and the LLaVA trained using our SAFE method on 5 commonly used benchmarks.}
\scalebox{0.75}{ 
\begin{tabular}{c|ccccccc}
\hline
Strategy &
  Method &
  POPE &
  VQA-v2 &
  VizWiz &
  GQA &
  MME &
  Average \\ \hline
\multirow{2}{*}{\begin{tabular}[c]{@{}c@{}}Training\\ Strategy1\end{tabular}} &
  Vanilla &
  85.63 &
  75.56 &
  47.54 &
  56.73 &
  69.63 &
  67.02 \\
 &
  \textbf{SAFE} &
  \textbf{85.70} &
  \textbf{75.58} &
  \textbf{52.09} &
  \textbf{57.20} &
  \textbf{70.69} &
  \textbf{68.25} \\ \hline
\multirow{2}{*}{\begin{tabular}[c]{@{}c@{}}Training\\ Strategy2\end{tabular}} & Vanilla & 85.56 & 75.72 & \textbf{48.96} & \textbf{56.94} & 72.24 & 67.88 \\
 &
  \textbf{SAFE} &
  \textbf{85.92} &
  \textbf{75.77} &
  \textbf{50.62} &
  \textbf{56.81} &
  \textbf{72.55} &
  \textbf{68.33} \\ \hline
\multirow{2}{*}{\begin{tabular}[c]{@{}c@{}}Training\\ Strategy3\end{tabular}} &
  Vanilla &
  85.53 &
  75.80 &
  49.64 &
  56.67 &
  73.03 &
  68.13 \\ 
 &
  \textbf{SAFE} &
  \textbf{86.19} &
  \textbf{75.80} &
  \textbf{50.34} &
  \textbf{56.82} &
  \textbf{73.45} &
  \textbf{68.52} \\ \hline
\end{tabular}
} \label{tab:llava}
\end{table}

Quantitatively, we considered five commonly used evaluation benchmarks, including MME \cite{fu2023mme}, VizWiz \cite{gurari2018vizwiz}, POPE \cite{li2023evaluating}, VQA-v2 \cite{goyal2017making}, and GQA \cite{hudson2019gqa}. As shown in Tab. \ref{tab:llava}, our method exhibits an improvement compared to the vanilla method to a certain extent. When using training strategy 1, the average performance of the SAFE method surpasses that of the Vanilla method by 1.23$\%$, whereas when using training strategy 3, the average performance of the SAFE method surpasses that of the Vanilla method by 0.39$\%$. The SAFE method does not introduce additional parameters and bears comparable training costs to the Vanilla method. Due to computational limitations, experiments were conducted on 4$\times$RTX 3090 GPUs rather than the recommended 8$\times$A100 GPUs. It's worth noting that, to ensure fair comparisons, the training settings for each experimental group remained consistent and adhered to official training recommendations. We have shared the trained models, training settings, and logs in this \href{https://drive.google.com/drive/folders/1GVmW3dW08NVlRaxf2CiI4eq9b33hL8y-?usp=sharing}{link}\footnote{https://drive.google.com/drive/folders/1GVmW3dW08NVlRaxf2CiI4eq9b \\ 33hL8y-?usp=sharing}. This paper focuses on exploring how to unleash the visual representation capability of the Contrastive Language-Image Pretraining (CLIP) model, aiming to achieve better performance in few-shot recognition tasks. The discussion in this subsection serves as a starting point for further research and is not the primary focus of this paper.

\section{Conclusion}\label{sec:conclusion}

In this paper, we provide novel insights into the obstacle factor in few-shot CLIP from the perspective of semantic attention and hypothesize that task-specific semantics matter in few-shot adaptation tasks.
Different from existing approaches that always freeze the entire CLIP parameters to avoid over-fitting and catastrophic forgetting caused by fine-tuning, we suggest fine-tuning CLIP's attention pooling layer to encourage the visual encoder to focus on task-specific semantics.
Additionally, we employ a residual blending during the inference between the outputs from the original attention pooling layer and the fine-tuned attention pooling layer to incorporate both the pre-trained prior knowledge and the few-shot knowledge, resulting in a significant performance gain in out-of-distribution scenarios. We name this lightweight and effective approach \textbf{S}emantic-\textbf{A}ware \textbf{F}in\textbf{E}-tuning (SAFE).
For the average accuracy across 11 benchmarks, SAFE achieves state-of-the-art performance that outperforms the second-best method by +1.51$\%$ under the 1-shot setting. 
We hope our findings can inspire new research into the internal mechanisms of CLIP to fully unleash its power for visual representation. 

% \section*{Acknowledgments}
% This work was supported by the Fundamental Research Funds for the Central Universities.

 {\small
\bibliographystyle{IEEEtranN}

\bibliography{newIEEEabrv}
}

%{\appendices
%\section*{Proof of the First Zonklar Equation}
%Appendix one text goes here.
% You can choose not to have a title for an appendix if you want by leaving the argument blank
%\section*{Proof of the Second Zonklar Equation}
%Appendix two text goes here.}

\vfill

\end{document}